\documentclass{article}
\usepackage{spconf,amsmath,amssymb,graphicx,floatrow}
\usepackage[sort&compress,square,comma,authoryear]{natbib}
\usepackage[group-separator={,}]{siunitx}

\usepackage{hyperref}
\usepackage[nameinlink,noabbrev]{cleveref}
\usepackage[T1]{fontenc}
\usepackage[utf8]{inputenc}

\hypersetup{
		breaklinks,
		linktoc=all,
		plainpages=false,
		unicode=true,
		colorlinks=false,
		hidelinks
}

\title{Neural Machine Translation with Characters and Hierarchical Encoding}

\name{\small Alexander Rosenberg Johansen~$^a$, Jonas Meinertz Hansen~$^a$, Elias Khazen Obeid~$^a$ Casper Kaae Sønderby~$^b$, Ole Winther~$^{a, b}$}

\address{\small $^a$ Department for Applied Mathematics and Computer Science, Technical University of Denmark (DTU), 2800 Lyngby, Denmark\\
         \small $^b$ Bioinformatics Centre, Department of Biology, University of Copenhagen, Copenhagen, Denmark}

\begin{document}
%
\maketitle
\begin{abstract}
\textbf{Most existing Neural Machine Translation models use groups of characters or whole words as their unit of input and output.
We propose a model with a hierarchical \texttt{char2word} encoder, that takes individual characters both as input and output.
We first argue that this hierarchical representation of the character encoder reduces computational complexity, and show that it improves translation performance.
Secondly, by qualitatively studying attention plots from the decoder we find that the model learns to compress common words into a single embedding whereas rare words, such as names and places, are represented character by character.}
\end{abstract}
%
%
\section{Introduction}
\label{sec:intro}

Neural Machine Translation (NMT) is the application of deep neural networks to translation of text. NMT is based on an end-to-end trainable algorithm that can learn to translate just by being presented with translated language pairs.
Despite being a relatively new approach, NMT has in recent years surpassed classical statistical machine translation models and now holds state-of-the-art results~\citep{wu2016google, LuongM16, ChungCB16}.

Early NMT models introduced by ~\citet{kalchbrenner2013recurrent, SutskeverVL14, ChoMBB14} are based on the \emph{encoder-decoder} network architecture.
Here the \emph{encoder} compresses an input sequence of variable length from the source language to a fixed-length vector representing the sentiment of the sentence.
The \emph{decoder}, takes the fixed-length representation as input and produces a variable length translation in the target language.
However, due to the fixed length representation the naïve encoder-decoder approach have limitations when translating longer sequences.

\begin{figure}[htb]
    \centering
    \includegraphics[width=\columnwidth]{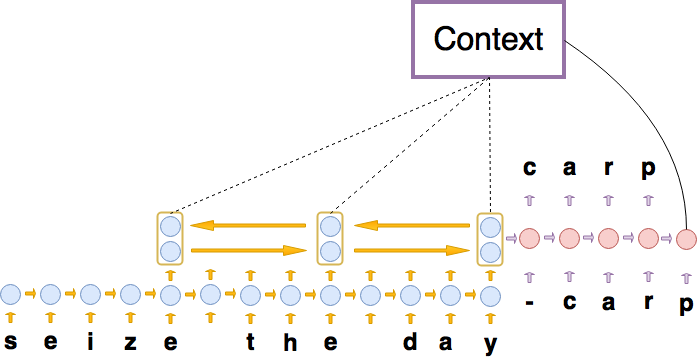}
    \caption{Our \texttt{char2word-to-char} model using hierarchical encoding in English and decoding one character at a time in Latin. ``-'' marks sequence start for decoding.}
    \label{fig:char2wordtochar}
\end{figure}

To overcome this shortcoming, the \emph{attention} mechanism proposed by \citet{BahdanauCB14} assists the decoder by learning to selectively attend to parts of the input sequence, which it deems most relevant for generating the next element in the output sequence and effectively reducing the distance from encoding to decoding.
This approach has made it possible for encoder-decoder models to produce high quality translations over longer sequences.
However, it suffers from the significant amount of computational power and memory needed to compute the relevance of every element of the input sequence for every element of the output sequence.

\begin{figure}[tb]
    \centering
    \includegraphics[width=\columnwidth]{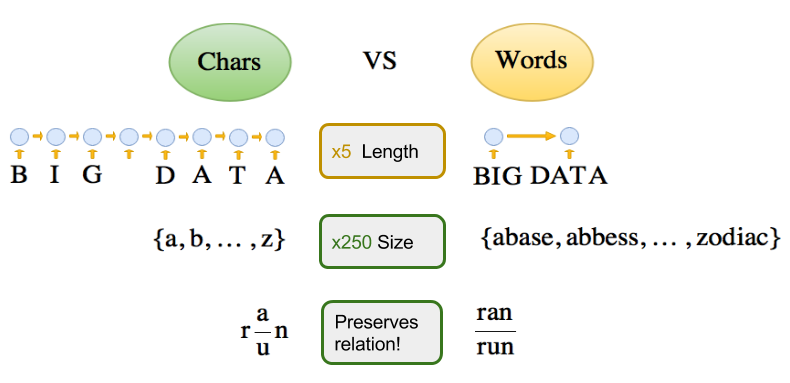}
    \caption{Challenges in \emph{encoder-decoder} models: Characters versus Words.}
    \label{fig:charvsword}
\end{figure}

For this reason, training the models on individual characters is not practical, and most current solutions instead use word segmentation~\citep{SutskeverVL14, BahdanauCB14} or multiple characters~\citep{SennrichHB15, wu2016google, schuster2012japanese} to represent sentences.
However, this high-level segmentation approach has a set of drawbacks; most Latin based languages have millions of words with the majority occurring rarely.
To handle this, current models use confined dictionaries with only the $k$ most common words with the remaining words being represented by a special \texttt{<UNK>}-token~\citep{SutskeverVL14, BahdanauCB14}.
As a result, names, places, and other rare words are typically translated as unknown by these models.
Further, when all words are represented as separate entities, the model has to learn how every word is related to one-another, which can be challenging for rare words even if they are obvious variations of more frequent words~\citep{LuongM16}.
\Cref{fig:charvsword} illustrates some of the challenges of characters versus words for the encoder-decoder model.

Two branches of methods to circumvent these drawbacks have been proposed.
The first branch is based on extending the current word-based encoder-decoder model to incorporate modules, such as dictionary look-up for out-of-dictionary words~\citep{LuongSLVZ14, JeanCMB14}.
The second branch, which we will investigate in this work, is moving towards smaller units of computation.

In this paper we demonstrate models that use a new \emph{char2word} mechanism (illustrated in~\cref{fig:char2wordtochar}) during encoding, which reduces long character-level input sequences to word-level representations.
This approach has the advantages of keeping a small alphabet of possible input values and preserving relations between words that are spelled similarly, while significantly reducing the number of elements that the attention mechanism needs to consider for each output element it generates.
Using this method the decoder's memory and computational requirements are reduced, making it feasible to train the models on long sequences of characters as input and output.
Thus avoiding the drawbacks of word based models described above.
And lastly, we give a qualitative analysis of attention plots produced by a character-level encoder-decoder model with and without the hierarchical encoding mechanism, char2word.
This shed light on how a character-level model uses attention, which might explain some of the success behind the BPE and hybrid models.

\section{Related work}
\label{sec:relatedwork}

Other approaches to circumventing the increase in sequence lengths while reducing the dictionary size have been proposed:
First, byte-pair Encoding (BPE)~\citep{SennrichHB15}, currently holding state-of-the-art in the WMT'14 English-to-French and English-to-German~\citep{wu2016google}, where the dictionary is a combination of the most common characters. In practice this means that most frequent words are encoded using fewer bits than less frequent words.
Secondly, a hybrid model~\citep{LuongM16} where a word encoder-decoder consults a character encoder-decoder when confronted with out-of-dictionary words.
Thirdly, pre-trained word-based encoder-decoder models with character input used for creating word embeddings~\citep{LingTDB15} have been shown to achieve similar results to word-based approaches.
As a last mention, character decoder with BPE encoder has shown to be end-to-end trained successfully~\citep{ChungCB16}.

\citet{wu2016google} provides a good summary and large-scale demonstration of many of the techniques that are known to work well for NMT and RNNs in general.
The RNN encoder-decoder with attention approach is used not only within machine translation, but can be regarded as a general architecture to extract information from a sequence and answer some type of question related to it~\citep{KumarISBEPOGS15}.

\section{Materials and Methods}
\label{sec:materialsandmethods}
First we give a brief description of the Neural Machine Translation model, afterwards we will go into detail of explaining our proposed architecture for character and word segmentation.
\subsection{Neural Machine Translation}
\label{ssec:neuralmachinetranslation}
From a probabilistic perspective, translation can be defined by maximizing the conditional probability of $\arg\max_yp(y|x)$, where $y_1, y_2, \dots, y_{T_y}$ is the target sequence and $x_1, x_2, \dots, x_{T_x}$ is the source sequence.
The conditional probability $p(y|x)$ is modelled by an encoder-decoder model where the encoder and the decoder are modelled by separate Recurrent Neural Networks (RNNs) and the whole model is trained end-to-end on a large parallel corpus.
The model uses memory based RNN variants, as they enable modelling of longer dependencies in the sequences~\citep{Hochreiter1997, ChoMGBSB14}.

The encoder part (input RNN) computes a set of hidden representations, $h_1, h_2, \dots, h_{T_x}$ based on the input
\begin{align}
\label{eq:defht}
h_t &= h(x_t, h_{t-1})
\end{align}
where $h$ is a RNN with memory cells, $h_t\in\mathbb{R}^{m_h}$ is a hidden state representation at time step $t$, with $m_h$ hidden units.

The decoder part (output RNN) then computes a context vector, $c_t$, based on the hidden representations from the encoder:
\begin{align}
\label{eq:ctxt}
c_t &= q({h_1, h_2, \dots, h_{T_x}}),
\end{align}
where $q$ is a function that takes a set of hidden representations and returns a context vector $c_t \in \mathbb{R}^{m_c}$ where $m_c$ number of context units.
For a decoder without attention, the value of $c_t$ is the same for all time steps.

Finally the decoder combines the previous predictions, $y_{<t}$, and the context vector, $c_t$, to predict the next unit (word, BPE, or character), such that it maximises the log conditional probability
\begin{align}
\label{eq:defg}
\log p(y \mid x) &= \sum_{t=1}^{T_y} \log p(y_t \mid y_{<t}, c),\\
p(y_t \mid y_{<t}, c_t) &= g(y_{t-1}, s_t, c_t),
\end{align}
where $g$ is a non-linear, potentially multi-layered function that outputs the probability of $y_t$, ande $s_t$ is the hidden state of the decoder RNN, such that
\begin{align}
\label{eq:defst}
s_t &= f(s_{t-1}, y_{t-1}, c_t).
\end{align}

We minimise the cross entropy loss averaged over all time steps with $n$ sized mini-batches and add $L^2$ regularisation, such that
\begin{equation*}
J = - \sum_{i=1}^{n} \log p(y_i \mid x_i) + \lambda  \biggl(\sum_{n'=1}^{N'}\theta^2_{n'}\biggr),
\end{equation*}
where $\lambda$ is a tune-able hyper-parameter, $N'$ is the number of non-bias weights and $\theta$ is the weights in the neural network.
\subsubsection{Attention}
\label{sssec:attention}
As motivated in \cref{sec:intro}, the attention mechanism can compute a new context vector $c_t$ for every time step by combining the hidden representations from the encoder as well as the previous hidden state, $s_{t-1}$, of the decoder
\begin{align}
c_t &= \sum_{j=1}^{T_x} a_{tj}h_j
\end{align}
where the weight parameter $a_{tj}$ of each annotation $h_j$ is computed as
\begin{align}
a_{tj} &= \frac{\exp(e_{tj})}{\sum_{k}^{T_x} \exp(e_{tk})}
\end{align}
and we have that
\begin{align}
\label{eq:defa}
e_{tj} &= a(s_{t-1}, h_j)
\end{align}
where $a_{tj}$ and $e_{tj}$ reflect the importance of $h_j$, w.r.t. the previous decoder state $s_{t-1}$.
The attention function, $a$, is a nonlinear, possibly multi-layered neural network.

The encoder-decoder and attention model is trained jointly to minimise the loss function.
\section{Our Model}
\label{ssec:ourmodel}
We propose two models: The \emph{char-to-char} NMT model and the \emph{char2word-to-char} NMT model.
Both models build on the encoder-decoder model with attention as defined in \cref{ssec:neuralmachinetranslation} and \cref{sssec:attention}.
Below we will give specific model definitions.
\subsection{The \texttt{char} encoder}
\label{sec:thecharencoder}
Our character-level encoder (referred to as the \texttt{char} encoder) is built upon a bi-directional RNN~\citep{schuster1997bidirectional}.
The encoder function, $h_t$, in \cref{eq:defht} becomes
\begin{align}
\label{eq:ht}
h_t = \begin{bmatrix}
h_f(Ex_t, \overrightarrow{h_{t-1}})  \\ 
h_b(Ex_t, \overleftarrow{h_{t+1}}), 
\end{bmatrix}
\end{align}
where $x_t \in \{0, 1\}^{m_x}$ is a one-hot encoded vector and $m_x$ is the amount of input classes, $E \in \mathbb{R}^{m_e\times m_x}$ is an embedding matrix with $m_e$ being the size of the embedding, $h_f$ and $h_b$ are RNN functions and $h$ is initialised as $h_0 = 0$.

The \texttt{char} encoder is illustrated with the yellow arrows and blue circles in \cref{fig:chartochar}.
\subsection{The \texttt{char2word} encoder}
\label{sec:thechar2wordencoder}
The character-to-word-level encoder (referred to as the \texttt{char\-2word} encoder) samples states from the forward pass of the \texttt{char} encoder defined in the above section.
The states it samples are based on the locations of spaces in the original text, resulting in a sequence of outputs from the \texttt{char} encoder that essentially represents the words from the text and acts as their embeddings.
We sample the indices from $\overrightarrow{h}$, such that
\begin{align}
h_t^{spaces} &= \overrightarrow{h}_{\varphi t},
\end{align}
where $\overrightarrow{h}_{t}$ is defined from above \cref{eq:ht} and $\varphi$ is an ordered list of indices defining spaces in the input sequence $x$.
Given $h^{spaces}$ \cref{eq:ht} is used with $h_t^{spaces}$ replacing $Ex_t$.

The \texttt{char2word} encoder is illustrated with the yellow arrows and blue circles in \cref{fig:char2wordtochar}.

A result of this ``downsampling'' by using spaces, the \texttt{char2word} encoder only has about a fifth of the hidden states the \texttt{char} encoder has.
As we described in the introduction, the computationally expensive part of the encoder-decoder with attention is the attention part.
By significantly reducing the encoder we could train the \texttt{char2word} encoder in half the time compared to the \texttt{char} encoder.
\subsection{The \texttt{char} decoder}
\label{sec:thechardecoder}
Our character-level decoder (referred to as the \texttt{char} decoder) works with characters as the smallest unit of computation and decodes one character at a time.
The decoder uses a RNN and the attention mechanism~\citep{BahdanauCB14} when decoding each character.

The new state in our decoder RNN, $s_t$, as defined in \cref{eq:defst} is computed as follows
\begin{align}
\label{eq:st}
s_t &= f(s_{t-1}, y_{t-1}, c_t), \\
s_t &= f(\begin{bmatrix}E'p_{t-1}\\c_t\end{bmatrix}, s_{i-1})\\
p_{t-1} &= \arg\max(y_{t-1}),
\end{align}
where $p_{t-1} \in {0, 1}^{k_p}$ is a one-hot encoded vector with $k_p$ being the amount of input classes, $E' \in \mathbb{R}^{m_e'\times m_p}$ is an embedding matrix with $m_e'$ being the size of the embedding, $f$ is a RNN function and $s$ is initialised as $s_0 = h_{T_x}$.
\subsubsection{Attention mechanism}
\label{sssec:attentionmechanism}
The attention model $a$ (defined in \cref{eq:a}) is used to compute the context $c_t$ for time step $t$, which is utilised by the decoder to perform variable length attention.
The attention function, $a$, was parametrized as
\begin{align}
\label{eq:a}
a(s_{t-1}, h_j) &= v_a^T\tanh(W_as_{t-1} + U_ah_j + b_a),
\end{align}
where $W_a\in \mathbb{R}^{m_s\times m_s}$, $U_a\in \mathbb{R}^{m_s\times m_h}$, $v_a\in \mathbb{R}^{m_h}$, $b_a\in \mathbb{R}^{m_h}$, $m_s$ is the amount of hidden units in the decoder and $m_h$ is the amount of hidden units in the encoder.
As $U_ah_j$ does not depend on $t$, we can pre-compute it in advance for optimisation purposes.
\subsubsection{Output function}
\label{sssec:outputfunction}
The output of the decoder $g(y_{t-1}, s_t, c_t)$ uses a linear combination of the current hidden state in the decoder, $s_t$, followed by a softmax function.
\begin{align}
\label{eq:g}
g(y_{t-1}, s_t, c_t) &= \frac{\exp(W_ys_t+b_y)}{\sum_{k=1}^K\exp(W_ys_i+b_y)},
\end{align}
where $W_y\in \mathbb{R}^{K\times m_s}$, $b_y\in \mathbb{R}^{K}$ and $K$ is the amount of output classes.

\begin{figure}[htb]
    \centering
    \includegraphics[width=\columnwidth]{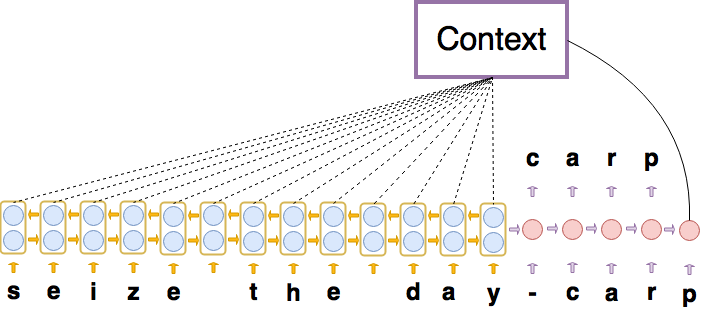}
    \caption{Our \texttt{char-to-char} model encoding and decoding a sentence from English to Latin on character level. ``-'' marks sequence start for decoding.}
    \label{fig:chartochar}
\end{figure}

We use the same decoder with attention for both the \texttt{char-to-char} and \texttt{char2word-to-char} model, which is illustrated in \cref{fig:char2wordtochar} and \cref{fig:chartochar}.
The main difference is that our \cref{fig:char2wordtochar} model has significantly lower amount of units to attend over.

\section{Experiments}
\label{sec:experiments}
All models were evaluated using the BLEU score\footnote{We used the \texttt{multi-bleu.perl} script from Moses (\url{https://github.com/moses-smt/mosesdecoder}).} \citep{Papineni2002}.
\subsection{Data and Preprocessing}
\label{ssec:data}
We trained our models on two different datasets of language pairs from the WMT'15: \texttt{En-De} (4.5M) and \texttt{De-En} (4.5M).
For validation we used the \texttt{newstest2013} and for testing we used \texttt{newstest2014} and \texttt{newstest2015}.

The data preprocessing applies is identical to \citet{ChungCB16} on \texttt{En-De} and \texttt{De-En} with the source sentence length set to 250 characters instead of 50 BPE units.
In short, that means;
We normalise punctuations and tokenise using Moses scripts\footnote{From Moses (\url{https://github.com/moses-smt/mosesdecoder}) using \texttt{normalize-punctuation.perl} and \texttt{tokenizer.perl}}.
We exclude all samples where the source sentence exceed 250 characters and the target sentence exceed 500 characters.
The source and target language has separate dictionaries, each containing the 300 most common characters.
Characters not in the dictionary is replaced with an unknown token.
\subsection{Training details}
\label{ssec:trainingdetails}
The model hyperparameters are listed in \cref{tab:char-to-char,tab:char2word-to-char}.
For the RNN functions in the encoder and the decoder we use gated recurrent units (GRU)~\citep{ChoMGBSB14}.
For training we use back-propagation with stochastic-gradient descent using the Adam optimiser~\citep{KingmaB14} with a learning rate of $\alpha = 0.001$.
For L2 regularization we set $\lambda=\num{1e-6}$.
In order to stabilise training and avoid exploding gradients, the norms of the gradients are clipped with a threshold of $1$ before updating the parameters.
All models are implemented using TensorFlow~\citep{tensorflow2015-whitepaper} and the code and details of the setup are available on GitHub\footnote{\url{https://github.com/Styrke/master-code}}.
\begin{table}[hbt]
\centering
\begin{tabular}{lc}
    \hline
    layer                      &  no. units  \\ \hline
    input alphabet size ($X$)  &  300        \\
    embedding sizes            &  256        \\
    char RNN (forward)         &  400        \\
    char RNN (backward)        &  400        \\
    attention                  &  300        \\
    char decoder               &  400        \\
    target alphabet size ($T$) &  300        \\ \hline
\end{tabular}
\caption{\label{tab:char-to-char}Hyperparameter values used for training the \texttt{char-to-char} model. Where $\Sigma_{src}$ and $\Sigma_{trg}$ represent the number of classes in the source and target languages, respectively.}
\end{table}
\begin{table}[hbt]
\centering
\begin{tabular}{lc}
    \hline
    layer                      &  no. units  \\ \hline
    input alphabet size ($X$)  &  300        \\
    embedding sizes            &  256        \\
    char RNN (forward)         &  400        \\
    spaces RNN (forward)       &  400        \\
    spaces RNN (backward)      &  400        \\
    attention                  &  300        \\
    char decoder               &  400        \\
    target alphabet size ($T$) &  300        \\ \hline
  \end{tabular}
  \caption{\label{tab:char2word-to-char}Hyperparameter values used for training the \texttt{char2word-to-char} model. Where $\Sigma_{src}$ and $\Sigma_{trg}$ represent the number of classes in the source and target languages, respectively.}
\end{table}
\subsubsection{Batch details}
\label{sssec:batchdetails}
\begin{figure}[htb]
    \centering
    \includegraphics[width=\columnwidth]{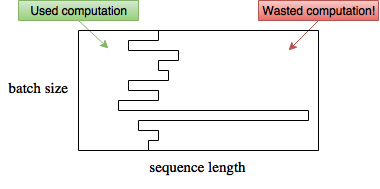}
    \caption{A regular batch with random samples.}
    \label{fig:dynamic_batching1}
\end{figure}

When training with batches, all sequences must be padded to match the longest sequence in the batch, and the recurrent layers must do the full set of computations for all samples and all timesteps, which can result in a lot of wasted resources~\citep{baiduDS} (see \cref{fig:dynamic_batching1}).
Training translation models is further complicated by the fact that source and target sentences, while correlated, may have different lengths, and it is necessary to consider both when constructing batches in order to utilize computation power and RAM optimally.

\begin{figure}[htb]
    \centering
    \includegraphics[width=\columnwidth]{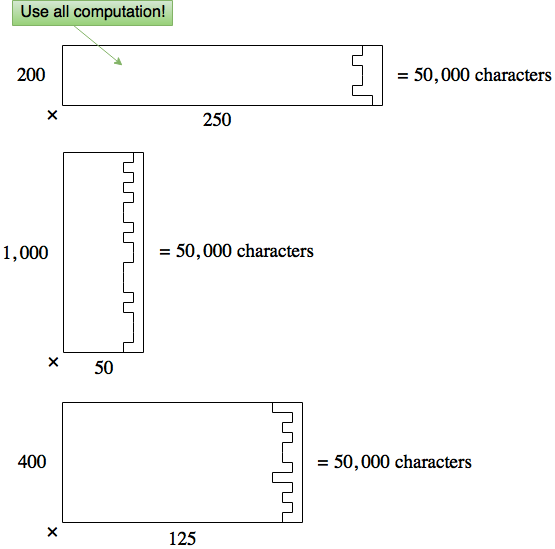}
    \caption{Our dynamic batches of variable batch size and sequence length.}
    \label{fig:dm6}
\end{figure}

To circumvent this issue, we start each epoch by shuffling all samples in the dataset and sorting them with a stable sorting algorithm according to both the source and target sentence lengths.
This ensures that any two samples in the dataset that have almost the same source and target sentence lengths are located close to each other in the sorted list while the exact order of samples varies between epochs.
To pack a batch we simply started adding samples from the sorted sample list to the batch, until we reached the maximal total allowed character threshold (which we set to $50{,}000$) for the full batch with padding after which we would start on a new batch.
Finally all the batches are fed in random order to the model for training until all samples have been trained on, and a new epoch begins.
\Cref{fig:dm6} illustrates what such dynamic batches might look like.

\subsection{Results}
\label{ssec:results}
\subsubsection{Quantitative}
\label{sssec:quantitative}
The quantitative results of our models are illustrated in \cref{tab:results-wmt}.
Notice that the \texttt{char2word-to-char} model outperforms the \texttt{char-to-char} model on all datasets (average $1.28$ BLEU performance increase).
This could be an indication that either having hierarchical, word-like, representations on the encoder or simply the fact that the encoder was significantly smaller, helps in NMT when using a character decoder with attention.
\begin{table*}[htb]
    \centering
      \begin{tabular}{llccc}
          \hline
                                                        &           & \emph{validation set} & \multicolumn{2}{c}{\emph{test sets}} \\
          Model                                         & Language  & newstest2013          & newstest2014   & newstest2015        \\ \hline
          \texttt{char-to-char}                         & De--En    & 18.89                 & 17.97          & 18.04               \\ 
          \texttt{char2word-to-char}                    & De--En    & \textbf{20.15}        & \textbf{19.03} & \textbf{19.90}      \\ \hline
          \texttt{char-to-char}                         & En--De    & 15.32                 & 14.15          & 16.11               \\
          \texttt{char2word-to-char}                    & En--De    & \textbf{16.78}        & \textbf{15.04} & \textbf{17.43}      \\ \hline
      \end{tabular}
    \caption{Results: WMT'15, \emph{newstest2013} was used as validation set, \emph{newstest2014} and \emph{newstest2015} were used as test sets.
      The results with bold indicates the best results on that dataset.}
    \label{tab:results-wmt}
\end{table*}
\subsubsection{Qualitative}
\label{sssec:qualitative}
\begin{figure*}[htb]
    \centering
    \includegraphics[width=\columnwidth]{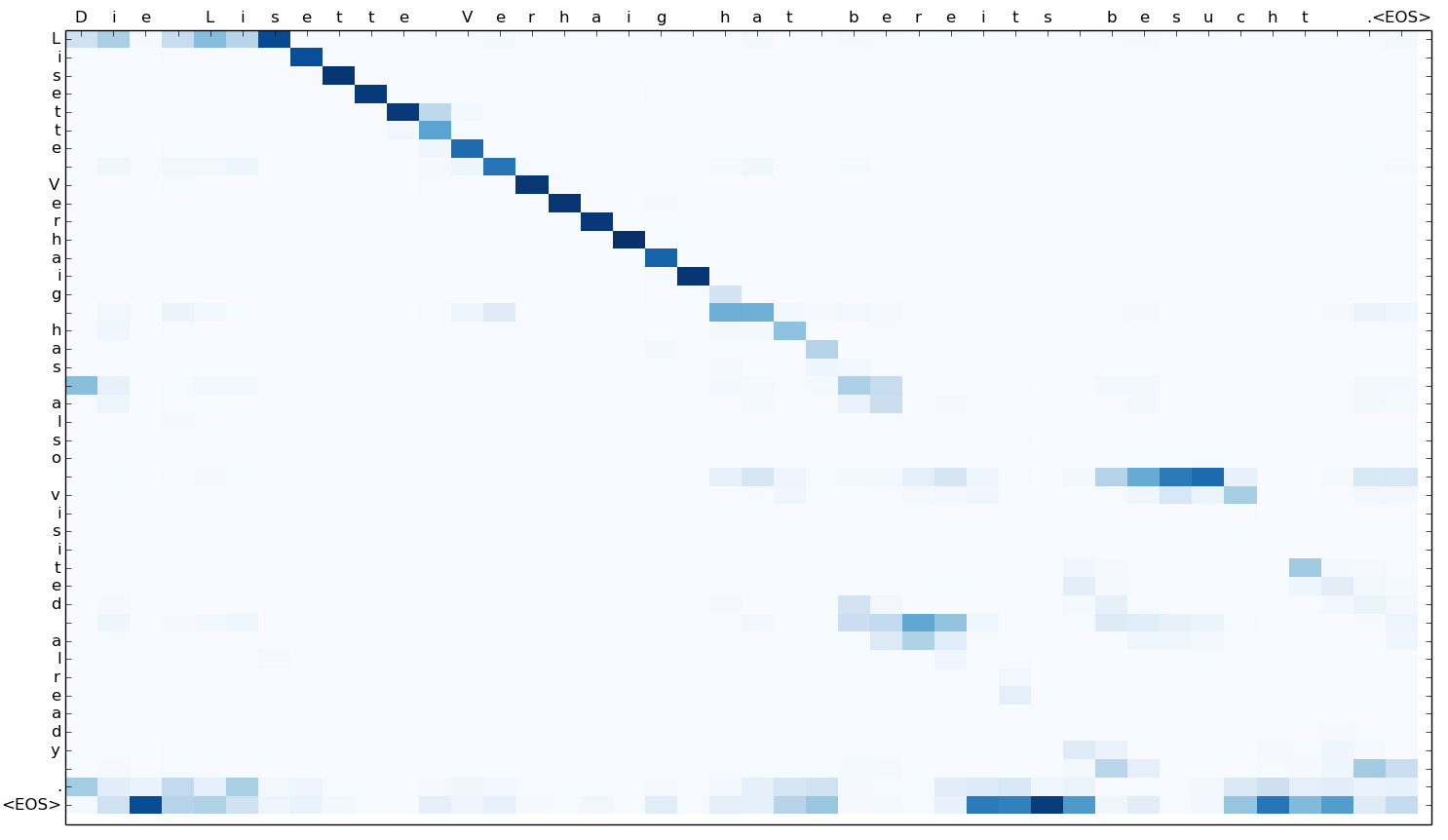}
    \caption{Attention plot of our \texttt{char-to-char} model encoding and decoding a sentence from English to German.}
    \label{fig:chartocharattention}
\end{figure*}

\begin{figure*}[htb]
    \centering
    \includegraphics[width=\columnwidth]{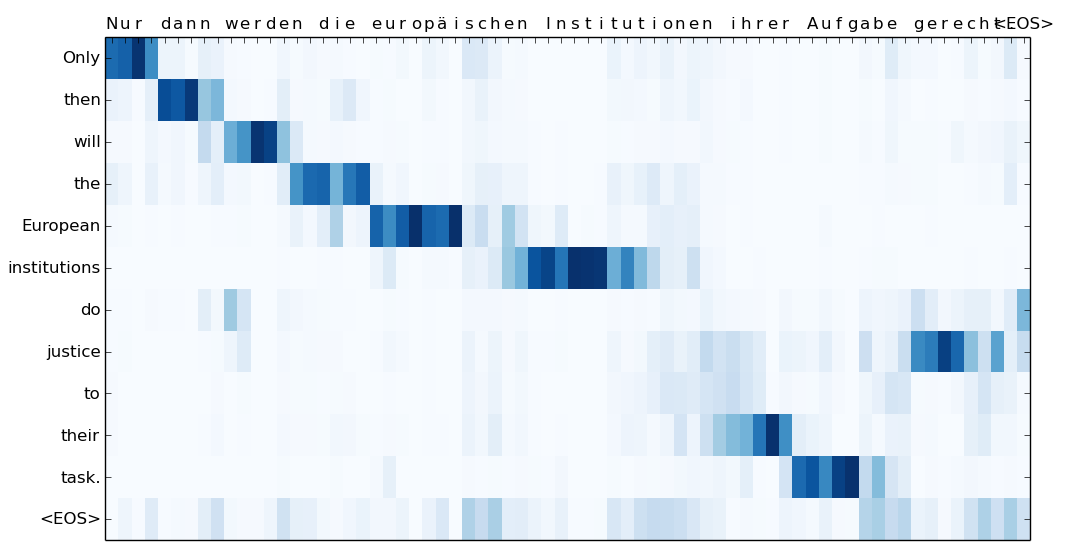}
    \caption{Attention plot of our \texttt{char2word-to-char} model encoding and decoding a sentence from English to German.}
    \label{fig:char2wordtocharattention}
\end{figure*}

Plotting the weights of $a_{tj}$ (defined at \cref{eq:a}) is popular in NMT research, as these gives an indication of where the model found relevant information while decoding.
We have provided plots of both our \texttt{char-to-char}- and \texttt{char2word-to-char} models in \cref{fig:chartocharattention,fig:char2wordtocharattention}.
The more intense the blue colour, the higher the values of $a_{tj}$ at that point.
Notice that each column corresponds to the decoding of a single unit, resulting in each column summing to $1$.

The char-to-char attention plot, attending over every character, interestingly indicates that words that would normally be considered out-of-dictionary (see \emph{Lisette Verhaig} in \cref{fig:chartocharattention}) are translated character by character-by-character, whereas common words are attended at the end/start of each word~\footnote{As we use a bi-directional RNN, full information will be available at both the end and start of a word} to use as a single embedding.
This observation might explain why using hierarchical encoding improves performance.
BPE based models and the hybrid word-char model by \citet{LuongM16} effectively works in the same manner, when translating common words BPE- and hybrid word-char models will work on a word level, whereas with rare words the BPE will work with sub-parts of the word (maybe even characters) and the hybrid approach will use character representations.

The \texttt{char2word-to-char} attention plot has words, or character-made embeddings of words, to perform attention over.
The attention plot seems very similar to the BPE-to-Char plot proposed by \citet{ChungCB16}.
This might indicate that it is possible to imitate lexeme (word) based models using smaller dictionaries and preserving relationship between words.

\section{Conclusion}
We have proposed a pure character based encoder-decoder model with attention using a hierarchical encoding.
We find that the hierarchical encoding, using our newly proposed char2word encoding mechanism, improves the the BLEU score by an average of $1.28$ compared to models using a standard character encoder.

Qualitatively, we find that the attention of a character model without hierarchical encoding learns to make hierarchical representations even without being explicitly told to do so, by switching between word and character embeddings for common and rare words.
This observation is in line with current research on Byte-Pair-Encoding- and hybrid word-character models, as these models uses word like embeddings for common words and sub-words or characters for rare words.

Furthermore, qualitatively we find that our hierarchical encoding finds lexemes in the source sentence when decoding similarly to current models with much larger dictionaries using Byte-Pair-Encoding.

\bibliographystyle{plainnat}
\bibliography{refs}

\begin{thebibliography}{20}
\providecommand{\natexlab}[1]{#1}
\providecommand{\url}[1]{\texttt{#1}}
\expandafter\ifx\csname urlstyle\endcsname\relax
  \providecommand{\doi}[1]{doi: #1}\else
  \providecommand{\doi}{doi: \begingroup \urlstyle{rm}\Url}\fi

\bibitem[Abadi et~al.(2016)Abadi, Agarwal, Barham, Brevdo, Chen, Citro,
  Corrado, Davis, Dean, Devin, Ghemawat, Goodfellow, Harp, Irving, Isard, Jia,
  J{\'{o}}zefowicz, Kaiser, Kudlur, Levenberg, Man{\'{e}}, Monga, Moore,
  Murray, Olah, Schuster, Shlens, Steiner, Sutskever, Talwar, Tucker,
  Vanhoucke, Vasudevan, Vi{\'{e}}gas, Vinyals, Warden, Wattenberg, Wicke, Yu,
  and Zheng]{tensorflow2015-whitepaper}
Mart{\'{\i}}n Abadi, Ashish Agarwal, Paul Barham, Eugene Brevdo, Zhifeng Chen,
  Craig Citro, Gregory~S. Corrado, Andy Davis, Jeffrey Dean, Matthieu Devin,
  Sanjay Ghemawat, Ian~J. Goodfellow, Andrew Harp, Geoffrey Irving, Michael
  Isard, Yangqing Jia, Rafal J{\'{o}}zefowicz, Lukasz Kaiser, Manjunath Kudlur,
  Josh Levenberg, Dan Man{\'{e}}, Rajat Monga, Sherry Moore, Derek~Gordon
  Murray, Chris Olah, Mike Schuster, Jonathon Shlens, Benoit Steiner, Ilya
  Sutskever, Kunal Talwar, Paul~A. Tucker, Vincent Vanhoucke, Vijay Vasudevan,
  Fernanda~B. Vi{\'{e}}gas, Oriol Vinyals, Pete Warden, Martin Wattenberg,
  Martin Wicke, Yuan Yu, and Xiaoqiang Zheng.
\newblock Tensorflow: Large-scale machine learning on heterogeneous distributed
  systems.
\newblock \emph{CoRR}, abs/1603.04467, 2016.
\newblock URL \url{https://arxiv.org/abs/1603.04467}.
\newblock Software available from tensorflow.org.

\bibitem[Bahdanau et~al.(2014)Bahdanau, Cho, and Bengio]{BahdanauCB14}
Dzmitry Bahdanau, Kyunghyun Cho, and Yoshua Bengio.
\newblock Neural machine translation by jointly learning to align and
  translate.
\newblock \emph{CoRR}, abs/1409.0473, 2014.
\newblock URL \url{https://arxiv.org/abs/1409.0473}.

\bibitem[Cho et~al.(2014{\natexlab{a}})Cho, van Merrienboer, Bahdanau, and
  Bengio]{ChoMBB14}
KyungHyun Cho, Bart van Merrienboer, Dzmitry Bahdanau, and Yoshua Bengio.
\newblock On the properties of neural machine translation: Encoder-decoder
  approaches.
\newblock \emph{CoRR}, abs/1409.1259, 2014{\natexlab{a}}.
\newblock URL \url{https://arxiv.org/abs/1409.1259}.

\bibitem[Cho et~al.(2014{\natexlab{b}})Cho, van Merrienboer,
  G{\"{u}}l{\c{c}}ehre, Bougares, Schwenk, and Bengio]{ChoMGBSB14}
Kyunghyun Cho, Bart van Merrienboer, {\c{C}}aglar G{\"{u}}l{\c{c}}ehre, Fethi
  Bougares, Holger Schwenk, and Yoshua Bengio.
\newblock Learning phrase representations using {RNN} encoder-decoder for
  statistical machine translation.
\newblock \emph{CoRR}, abs/1406.1078, 2014{\natexlab{b}}.
\newblock URL \url{https://arxiv.org/abs/1406.1078}.

\bibitem[Chung et~al.(2016)Chung, Cho, and Bengio]{ChungCB16}
Junyoung Chung, Kyunghyun Cho, and Yoshua Bengio.
\newblock A character-level decoder without explicit segmentation for neural
  machine translation.
\newblock \emph{CoRR}, abs/1603.06147, 2016.
\newblock URL \url{https://arxiv.org/abs/1603.06147}.

\bibitem[Hannun et~al.(2014)Hannun, Case, Casper, Catanzaro, Diamos, Elsen,
  Prenger, Satheesh, Sengupta, Coates, and Ng]{baiduDS}
Awni~Y. Hannun, Carl Case, Jared Casper, Bryan Catanzaro, Greg Diamos, Erich
  Elsen, Ryan Prenger, Sanjeev Satheesh, Shubho Sengupta, Adam Coates, and
  Andrew~Y. Ng.
\newblock Deep speech: Scaling up end-to-end speech recognition.
\newblock \emph{CoRR}, abs/1412.5567, 2014.
\newblock URL \url{https://arxiv.org/abs/1412.5567}.

\bibitem[Hochreiter and Schmidhuber(1997)]{Hochreiter1997}
Sepp Hochreiter and J\"{u}rgen Schmidhuber.
\newblock Long short-term memory.
\newblock \emph{Neural Comput.}, 9\penalty0 (8):\penalty0 1735--1780, November
  1997.
\newblock ISSN 0899-7667.
\newblock \doi{10.1162/neco.1997.9.8.1735}.
\newblock URL \url{http://dx.doi.org/10.1162/neco.1997.9.8.1735}.

\bibitem[Jean et~al.(2014)Jean, Cho, Memisevic, and Bengio]{JeanCMB14}
S{\'{e}}bastien Jean, Kyunghyun Cho, Roland Memisevic, and Yoshua Bengio.
\newblock On using very large target vocabulary for neural machine translation.
\newblock \emph{CoRR}, abs/1412.2007, 2014.
\newblock URL \url{https://arxiv.org/abs/1412.2007}.

\bibitem[Kalchbrenner and Blunsom(2013)]{kalchbrenner2013recurrent}
Nal Kalchbrenner and Phil Blunsom.
\newblock Recurrent continuous translation models.
\newblock In \emph{EMNLP}, volume~3, page 413, Seattle, October 2013.
  Association for Computational Linguistics.

\bibitem[Kingma and Ba(2014)]{KingmaB14}
Diederik~P. Kingma and Jimmy Ba.
\newblock Adam: {A} method for stochastic optimization.
\newblock \emph{CoRR}, abs/1412.6980, 2014.
\newblock URL \url{https://arxiv.org/abs/1412.6980}.

\bibitem[Kumar et~al.(2015)Kumar, Irsoy, Su, Bradbury, English, Pierce,
  Ondruska, Gulrajani, and Socher]{KumarISBEPOGS15}
Ankit Kumar, Ozan Irsoy, Jonathan Su, James Bradbury, Robert English, Brian
  Pierce, Peter Ondruska, Ishaan Gulrajani, and Richard Socher.
\newblock Ask me anything: Dynamic memory networks for natural language
  processing.
\newblock \emph{CoRR}, abs/1506.07285, 2015.
\newblock URL \url{https://arxiv.org/abs/1506.07285}.

\bibitem[Ling et~al.(2015)Ling, Trancoso, Dyer, and Black]{LingTDB15}
Wang Ling, Isabel Trancoso, Chris Dyer, and Alan~W. Black.
\newblock Character-based neural machine translation.
\newblock \emph{CoRR}, abs/1511.04586, 2015.
\newblock URL \url{https://arxiv.org/abs/1511.04586}.

\bibitem[Luong and Manning(2016)]{LuongM16}
Minh{-}Thang Luong and Christopher~D. Manning.
\newblock Achieving open vocabulary neural machine translation with hybrid
  word-character models.
\newblock \emph{CoRR}, abs/1604.00788, 2016.
\newblock URL \url{https://arxiv.org/abs/1604.00788}.

\bibitem[Luong et~al.(2014)Luong, Sutskever, Le, Vinyals, and
  Zaremba]{LuongSLVZ14}
Thang Luong, Ilya Sutskever, Quoc~V. Le, Oriol Vinyals, and Wojciech Zaremba.
\newblock Addressing the rare word problem in neural machine translation.
\newblock \emph{CoRR}, abs/1410.8206, 2014.
\newblock URL \url{https://arxiv.org/abs/1410.8206}.

\bibitem[Papineni et~al.(2002)Papineni, Roukos, Ward, and Zhu]{Papineni2002}
Kishore Papineni, Salim Roukos, Todd Ward, and Wei-Jing Zhu.
\newblock Bleu: A method for automatic evaluation of machine translation.
\newblock In \emph{Proceedings of the 40th Annual Meeting on Association for
  Computational Linguistics}, ACL '02, pages 311--318, Stroudsburg, PA, USA,
  2002. Association for Computational Linguistics.
\newblock \doi{10.3115/1073083.1073135}.
\newblock URL \url{http://dx.doi.org/10.3115/1073083.1073135}.

\bibitem[Schuster and Paliwal(1997)]{schuster1997bidirectional}
M.~Schuster and K.K. Paliwal.
\newblock Bidirectional recurrent neural networks.
\newblock \emph{Trans. Sig. Proc.}, 45\penalty0 (11):\penalty0 2673--2681,
  November 1997.
\newblock ISSN 1053-587X.
\newblock \doi{10.1109/78.650093}.
\newblock URL \url{http://dx.doi.org/10.1109/78.650093}.

\bibitem[Schuster and Nakajima(2012)]{schuster2012japanese}
Mike Schuster and Kaisuke Nakajima.
\newblock Japanese and korean voice search.
\newblock In \emph{2012 IEEE International Conference on Acoustics, Speech and
  Signal Processing (ICASSP)}, pages 5149--5152. IEEE, 2012.

\bibitem[Sennrich et~al.(2015)Sennrich, Haddow, and Birch]{SennrichHB15}
Rico Sennrich, Barry Haddow, and Alexandra Birch.
\newblock Neural machine translation of rare words with subword units.
\newblock \emph{CoRR}, abs/1508.07909, 2015.
\newblock URL \url{https://arxiv.org/abs/1508.07909}.

\bibitem[Sutskever et~al.(2014)Sutskever, Vinyals, and Le]{SutskeverVL14}
Ilya Sutskever, Oriol Vinyals, and Quoc~V. Le.
\newblock Sequence to sequence learning with neural networks.
\newblock \emph{CoRR}, abs/1409.3215, 2014.
\newblock URL \url{https://arxiv.org/abs/1409.3215}.

\bibitem[Wu et~al.(2016)Wu, Schuster, Chen, Le, Norouzi, Macherey, Krikun, Cao,
  Gao, Macherey, et~al.]{wu2016google}
Yonghui Wu, Mike Schuster, Zhifeng Chen, Quoc~V Le, Mohammad Norouzi, Wolfgang
  Macherey, Maxim Krikun, Yuan Cao, Qin Gao, Klaus Macherey, et~al.
\newblock Google's neural machine translation system: Bridging the gap between
  human and machine translation.
\newblock \emph{arXiv preprint arXiv:1609.08144}, 2016.
\newblock URL \url{https://arxiv.org/abs/1609.08144}.

\end{thebibliography}

\end{document}